\documentclass[letterpaper, 10 pt, conference]{ieeeconf}  % Comment this line out if you need a4paper

\usepackage{color}
\usepackage{graphicx}
\usepackage{booktabs}
\usepackage{multirow}
\usepackage{siunitx}
\usepackage{multicol}
\usepackage{leftidx}
\usepackage{amssymb}
\usepackage{amsmath}
\usepackage{amsfonts}
\usepackage{graphicx}
\usepackage{graphics}
\IEEEoverridecommandlockouts                              % This command is only needed if 
\makeatletter
\DeclareRobustCommand\onedot{\futurelet\@let@token\@onedot}
\def\@onedot{\ifx\@let@token.\else.\null\fi\xspace}

\makeatother

\title{\LARGE \bf
A Hierarchical Network for Diverse Trajectory Proposals
}

\author{ Sriram N. N.$^{1}$, Gourav Kumar$^{1}$, Abhay Singh$^{1}$, M. Siva Karthik$^{2}$, Saket Saurav$^{1}$\\  Brojeshwar Bhowmick$^{3}$ and  K. Madhava Krishna$^{1}$% <-this % stops a space
\thanks{$^{1}$ International Institute of Information Technology, Hyderabad, India}
\thanks{$^{2}$ University of Heidelberg, Germany}
\thanks{$^{3}$ TCS Research and Innovation Labs, Kolkata, India}
}

\begin{document}

\maketitle
\thispagestyle{empty}
\pagestyle{empty}

\begin{abstract}

Autonomous explorative robots frequently encounter scenarios where multiple future trajectories can be pursued. Often these are cases with multiple paths around an obstacle or trajectory options towards various frontiers. Humans in such situations can inherently perceive and reason about the surrounding environment to identify several possibilities of either manoeuvring around the obstacles or moving towards various frontiers. In this work, we propose a 2 stage Convolutional Neural Network architecture which mimics such an ability to map the perceived surroundings to multiple trajectories that a robot can choose to traverse. The first stage is a Trajectory Proposal Network which suggests diverse regions in the environment which can be occupied in the future. The second stage is a Trajectory Sampling network which provides a finegrained trajectory over the regions proposed by Trajectory Proposal Network. We evaluate our framework in diverse and complicated real life settings. For the outdoor case, we use the KITTI dataset and our own outdoor driving dataset. In the indoor setting, we use an autonomous drone to navigate various scenarios and also a ground robot which can explore the environment using the trajectories proposed by our framework. Our experiments suggest that the framework is able to develop a semantic understanding of the obstacles, open regions and identify diverse trajectories that a robot can traverse. 
 Our comparisons portray the performance gain of the proposed architecture over a diverse set of methods against which it is compared.

% Mention that we dont use goal points.
% Compares well to RRT ?
% Identifies openings, intersections etc.
% Comclude the abstract.

\end{abstract}
\section{Introduction}

Autonomous navigation requires the explorative ability to navigate by identifying diverse paths to multiple goal points by perceiving its environment. A simple instance is a case where an autonomous drone using SLAM \cite{Edge}  or using any other sensor based reconstruction \cite{mob1, mob2} can manoeuvre between obstacles by going around towards their right or left side. Similarly, in an indoor corridor intersection, it can proceed straight, right or left. The autonomous robot has to identify various possible paths and goal points that it can pursue in any given setup. Such a task is easy for humans where they can map the scene configuration to identify multiple traversable areas, goal points and also a fine grained path. One such use case in autonomous vehicle setting is where you know the rough direction of travel but not the actual goal point. A diverse path prediction is required in this case to determine the best plan of action. Some of the key applications can include but not limited to:

%In this work, we propose a Convolutional Neural Network framework which mimics such capabilities to implicitly identify multiple goal points, traversable areas and finally a diverse set of trajectories towards various goal points. 

\begin{figure}[!ht]
% \begin{center}
\includegraphics[width=9cm, height=6.7cm]{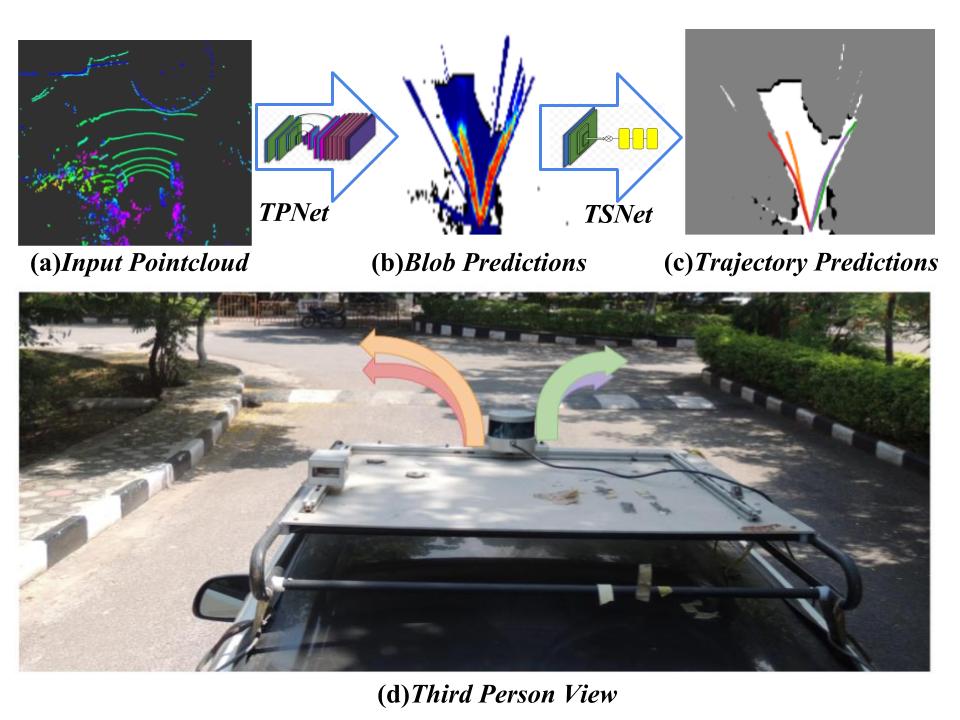}
\label{fig:teaser} 
\caption{\textbf{Trajectory Proposal:}(a) Bird's-eye-view of the pointcloud data generated at a road bifurcation. This pointcloud data is converted into a 2D occupancy grid and passed as an input to \emph{TPNet} (b) Proposed trajectories in the form of probability values for each pixel which is passed to \emph{TSNet} (c) Sampled trajectory output. (d) Third-person-view of the driving scenario illustrating proposed trajectories with respect to the car.}
\vspace{-0.5cm}
\end{figure}

\begin{figure}[t!]
\centering
\includegraphics[width=4cm,height=3cm]{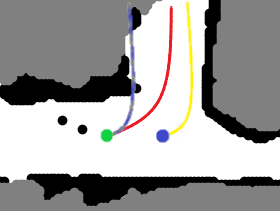}
\caption{Planning failures due to localization based on GPS. Green dot represents the true location of the ego vehicle while the blue dot represents the erroneous location given by GPS. Hence, planning a trajectory based on the GPS location (yellow line) and executing the trajectory on the true location might not be feasible (blue line). Our proposed method gives a trajectory based on the scene structure (red line) which further can help in correcting these errors. }
\label{fig:gps_error}
  \vspace{-0.8cm}
\end{figure}

\begin{itemize}
    \item In case of rerouting alternate trajectories might be required due to sudden unexpected changes in the environment.
    \item It also opens up the possibility of reaching the destination in multiple possible ways.
    % \item Mapping and localization on an outdoor terrain with a ground robot or an autonomous vehicle.
    \item It can be useful in overcoming GPS errors. The networks output can  help in guiding the vehicle precisely to take turns thus overcoming the state estimation errors that results in early or late turns near an intersection as shown in figure \ref{fig:gps_error}.
\end{itemize}

In this work, we present a Convolutional Neural Network framework which  precludes the need for explicit determination of candidate waypoints as it learns a direct mapping from intermediate semantic representation to candidate trajectories as shown in fig. \ref{fig:teaser}. Our framework which  consists of two stages. The first is a Trajectory Proposal Network (TPNet), an encoder-decoder style network, which uses the occupancy map and robot's past trajectory as input and produces multiple traversable areas by inherently discovering the waypoints possible in the scene.
%The occupancy map indicates the obstacles, free space and the unexplored areas in the environment. 
This is achieved through our novel supervision based on multiple choice learning which encourages the TPNet to identify various waypoints and propose diverse traversable areas for a given scenario. The second stage of our framework, the Trajectory Sampler Network (TSNet) is a Long Short-Term Memory (LSTM) based network which uses the proposals by TPNet and samples precise trajectories values through those proposals which can lead the robot to various goal points. This alleviates the need to explicitly sample various goal points and compute trajectories towards them. Further, our framework generalizes well when trained on outdoor data and evaluated on indoor data and vice-versa. This is because we use input data which is occupancy based instead of appearance based.
% Also, since our network relies on locally perceived data to generate candidate proposals it circumvents problems of overshoot or undershoot of waypoints at intersections due to GPS errors and could in turn correct the robot's waypoints locally as shown in figure \ref{fig:gps_error}.

%Since our network relies on perceptual input to generate candidate proposals it circumvents problems of overshoot or undershoot of waypoints at intersections due to GPS errors and can in turn correct the robot's waypoints locally.
%Our network precludes the need for explicit determination of candidate waypoints as it learns a direct mapping from intermediate semantic representation to candidate trajectories. 

To evaluate the efficacy of our network, we perform extensive experimentation in outdoor scenes using the KITTI \cite{kitti_odom} scenes and our own outdoor driving scenes. Additionally, we demonstrate results in an indoor setting in a laboratory environment with varied complexity in scene structure.  To quantitatively evaluate our network, we compare our results to traditional path planners like RRT-star \cite{rrt_star}, BIT-star \cite{bit_star} etc in terms of similarity between our paths and theirs for a particular goal point and the time taken to estimate the paths. To this end, our contributions are the following :

\begin{itemize}
	\item We present a novel CNN framework which is able to map the perceived surroundings to a diverse set of trajectories towards inherently inferred goal points. Such abilities are very relevant in autonomous navigation settings where alternate feasible trajectories are needed due sudden changes in the motions of surrounding vehicles.
	\item Our proposed supervision strategy to train the TPNet based on Multiple Choice Learning enables the network to identify diverse traversable areas which are further used to predict trajectories towards various goal points.
	\item Our framework easily generalizes to new scenarios due to the versatility of the intermediate representation using occupancy maps. For instance, a network trained on LIDAR data for outdoors works for a monocular or indoor depth maps obtained from drones.
	\item We demonstrate results on the KITTI dataset which uses Velodyne 64 , our campus dataset with Velodyne 16 and also  scenarios like indoor drone  which uses an RGBD camera.
	%Also summarize the improvements due to our method over classical methods in terms of constant time outputs despite varying number of choices possible.
\end{itemize}

We demonstrate that our network performs better than classical trajectory prediction methods in terms of time taken as our network predicts trajectories in constant time despite increasing the number of possible choices. To the best of our knowledge, there has not been prior literature that addresses the problem in such an end to end fashion.

% Things to emphasize :
% Mention explicitly that it does not need goal point inputs. 
% Maps semantics to traversible paths
% Training advantage with multiple choice learning
% About the GPS thingy! Overshoot and undershoot
%Using the proposed trajectories as navigation priors in order to improve and enhance the planned trajectories making the planner less prone to localization errors. Unlike previous methods[] in which the scene information is not considered or is assumed that the perception has been solved our proposed method fuses the scene information to predict trajectories. 
 
\section{Related Work}
The problem of identifying multiple driving/navigation options for a given perceptual input has not been widely studied in literature. While not exactly a trajectory planning problem, classical planners such as \cite{rrt}, \cite{rrt_star} would pose this problem as one of hypothesizing/sampling multiple goal locations followed by computation of trajectories to such candidate goal locations.  Evidently as the diversity of navigation options increase the computational time for generating all possible candidate trajectories increases. In contrast in this paper the end to end learning algorithm provides for constant time outputs despite increasing number of navigation options or driver intents as the case may be. The task of trajectory estimation and planning has been attempted before using neural networks. Glasius et al. \cite{glasius} presented an approach based on Hopfield Neural Networks for generating paths in dynamic environments. Yang et al. \cite{yang} presented a computationally efficient neural architecture for real time navigation of robots in dynamic environments. The closest formulations in the literature are those that predict driver intents around an intersection \cite{LSTM-MDN}, \cite{intention_prediction_intersection}, \cite{roundabout}, however such methods are restrictive in that they rely on explicit knowledge of the intersection and do not scale to situations beyond intersections. There are a number of methods that map perceptual inputs to continuous space control actions \cite{abbeele2e} , \cite{dronet} in an end to end framework. However extensions of such formulations to predict multiple trajectory proposals have not appeared in literature thereby placing the current algorithm as distinct in the context of existing works. On the other hand, robots would some times want to explore the possibilities of where they could traverse in a given scene. The current method works on intermediate representations obtained by a variety of robotic agents from diverse sensors such as stereo cameras, RGBD sensors and LIDAR to generate candidate navigating options (trajectories). We present an approach which generalizes to various scenarios like autonomous drones navigating in indoor and outdoor environments, ground robots exploring indoor scenes and also autonomous cars in outdoor scenes.

\section{Method}

\begin{figure*}[h!]
\centering
\includegraphics[width=18cm]{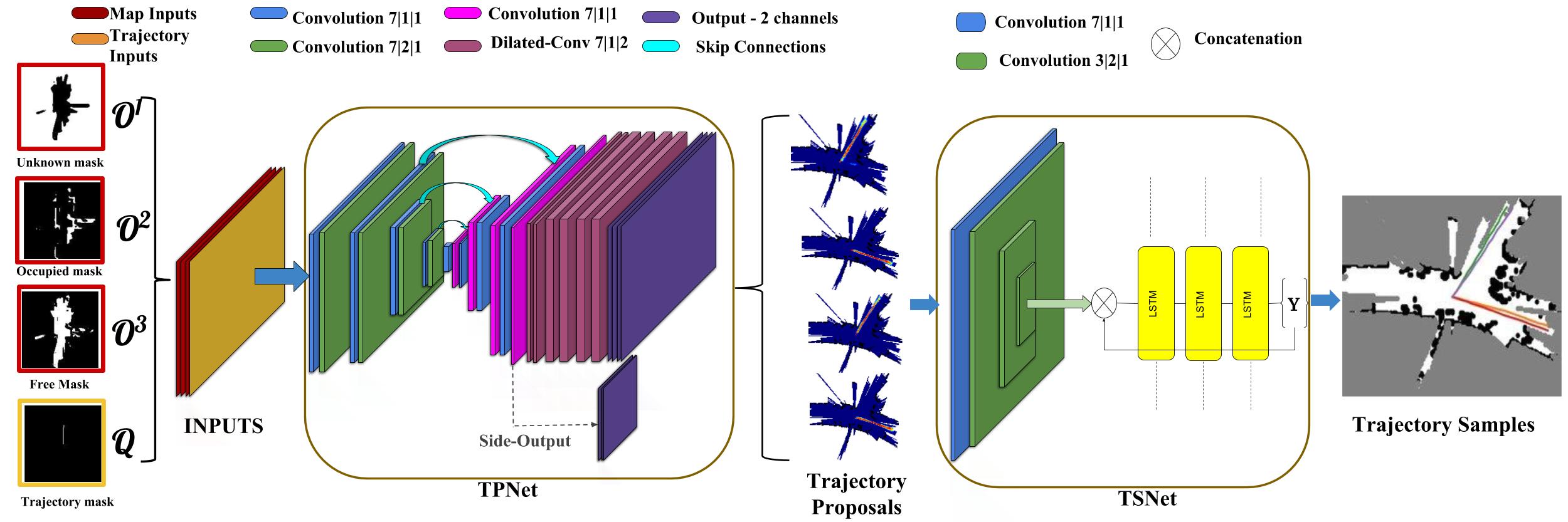}
\caption{\emph{Pipeline(Forward Pass)}: Given the stacked binary masks corresponding to free, occluded, unknown regions and past trajectory points respectively as inputs, the  \textbf{TPNet} predicts four \emph{Trajectory Proposals} as a set of probability values corresponding to each grid pixels. Each of these \emph{Trajectory Proposals} are then fed as inputs to \textbf{TSNet} which produces \emph{Trajectory Samples} as a set of waypoints.}
\label{overall_pipeline}
\end{figure*}

In this section, we describe the details of our architecture and formally define the problem and our approach below. The overall pipeline of the proposed method is illustrated in the figure \ref{overall_pipeline}.

%The motivation is to get a set of plausible future trajectories positions that the vehicle might occupy.

%Using the proposed trajectories as navigation priors in order to improve and enhance the planned trajectories making the planner less prone to localization errors. Unlike previous methods[] in which the scene information is not considered or is assumed that the perception has been solved our proposed method fuses the scene information to predict trajectories. 

\subsection{Inputs}
We use 2 types of inputs to the Trajectory Proposal Network. The first input is $\mathcal{O}$ which represents the occupied space ($\mathcal{O}^1$), free space ($\mathcal{O}^2$) and the unknown/unexplored space ($\mathcal{O}^3$) in the perceivable area of the robot. These are computed using the lidar based point cloud and identifying all the potential obstacles within a certain range, the free space and the unexplored area. $\mathcal{O}$ is obtained by projecting the registered point cloud data to 2D grid in birds-eye view and is the stack of mutually exclusive binary masks {$\mathcal{O}^1, \mathcal{O}^2, \mathcal{O}^3$} each of dimensions $h\times w$. Hence, ({\it $\mathcal{O}_{xy}^i$=1})$\Rightarrow$ ({\it $\mathcal{O}_{xy}^{j\neq i}$ = 0}) $\forall (x,y) \in (h\times w)$. The second input type ${\mathcal{Q}_{h\times w\times 1}}$ specifies the track history of the ego vehicle in the from of grid with {\it $\mathcal{Q}_{xy}$} representing whether the {\it $xy^{th}$} grid cell was occupied by the ego vehicle in the past. Effectively, the input to the network is ${\mathcal{I}_{h\times w\times 4}}$ =  $\{\mathcal{O,Q}\}$. $\mathcal{I}$ has ego-centric vehicle information. The vehicle’s track histories are appropriately transformed to the ego vehicles coordinate frame and then discretized to form a grid.

%The input to the network is, \textbf{\it I$_{h\times w\times 5}$} =  {\{\it {\bf O},Q,L\}} where, 
%{\bf \it \bf O$_{h\times w\times 3}$} is stack of three mutually exclusive binary masks {\it O$^i$} of size {\it $h \times w$ } representing occupied ({\it O$^1$}), free space ({\it O$^2$}) and unknowns ({\it O$^3$}) respectively. Here, {\it h,w }$\in \mathbb{R}$ are the grid dimensions. 
%The binary mask can be represented by each cell in {\it O$_{xy}^i \in \{0,1\}$}. Also the three channels are mutually exclusive, 
%{\it Q$_{h\times w\times 1}$} specifies the track history of the ego vehicle in the from of grid with {\it Q$_{xy}$} representing whether the {\it $xy^{th}$} grid cell was occupied in the past. {\it L$_{h\times w\times 1}$} is used to input processed lane information in the form of a grid whenever available. 
%The input {\it \bf I} has ego-centric vehicle information. {\it \bf O} is formed by projecting the registered point cloud data in 2D grid form and then decomposing it into three mutually exclusive binary masks. The vehicle’s location is always in the centre of the grid map. The vehicle’s track histories are appropriately transformed to the ego vehicles coordinate frame and then discretised to form a grid.

\subsection{Training Data}
To obtain training ground truth for our framework, we use an RRTstar \cite{rrt_star} based planner from Open Motion Planning Library \emph{OMPL} \cite{ompl}. For each of the training scenario in $\mathcal{I}_{train}$, multiple goal points are chosen in a manner that it ensures all dominant choices of motion. 
% we estimate multiple frontier points to which the robot can proceed. 
Further, we obtain trajectories to each of these goal points using RRTstar and discretise them over the grid. Such an approach enables us to create large amounts of training data by using lidar data and RRT based planner without any need for annotation. To this end, for a scene $I_t \in \mathcal{I}_{train}$,   $\mathcal{P}_t = \{P_i\}$ represents the diverse trajectories possible where $\textit{i}$ varies from $1$ to $N$ corresponding to each scene.

%We require that our framework produces diverse trajectories in the possible navigation space. \\ \\
%\textbf{P} = \{\textit{p$_i$}\} represents the supervised vehicle trajectories where {\it i} varies from 1 to N depending on the number of different trajectories possible given the input data. 
%We use an RRTstar [cite] based planner from \emph{OMPL} \cite{ompl} for trajectory supervision and these are discretised and 4 connected in the form of a grid. 

\subsection{Trajectory Proposal Network}
The Trajectory Proposal Network is the first module in our framework which proposes diverse multi-modal areas from which trajectories can further be sampled by the Trajectory Sampling Network. These are regions in the open space which could be traversed by the robot.\\
\subsubsection{Architecture}
 We use a CNN with an Encoder-Decoder style architecture with skip connections. All the convolutions in the network are Dilated. The output of the network is $\mathcal{R}$ = $\{R_k\}$ where $k$ is the number of diverse regions we want the network to predict. $R_k$ has 2 channels which indicate the probability of each pixel in the channel belonging to traversable region or not. Effectively, the network has $k \times 2$ channels and through various experiments we chose $k$ = $4$ in our setup as it was able to capture all possible trajectories in most of the situations.
% \siva{mention more network details here.}\\ 

% MENTION ABOUT THE NETWORK SPREADING ITS BETS BECAUSE IT HAS MULTIPLE OUTPUTS
\subsubsection{Training}
%We train our network using three kinds of supervision. Firstly, 
We train the TPNet in a supervised fashion using $<I_t, P_t>$ pairs. For each training iteration, we randomly pick $I_t$ and one of the groundtruth trajectories $P_i$ from the pool $\mathcal{P}_t$. Our first loss which we call as \textit{Trajectory Diversity Loss}, $\mathcal{L}_{td}$ defined in eq. \ref{eq:l_td} encourages the network to predict diverse proposals through its output $\mathcal{R}$. To compute this, we evaluate the weighted cross entropy loss between each of the trajectory outputs $R_k$ in $\mathcal{R}$ with the ground truth trajectory $P_i$ and choose the minimum of these losses, which is the best possible proposal generated. This is inspired from the Multiple Choice Learning framework presented in \cite{mcl}. Such a loss encourages the network to spread its bets on various proposals in its multiple output layers.

%\begin{equation}
%    \mathcal{L}_{td} = \sum_{j} \frac{1}{N_B}(\sum_{i=1}^{N_B} \min_{k} \hat{\mathcal{T}}_j^k)
%\label{eq:l_td}    
%\end{equation}
%\begin{equation}
%    \hat{\mathcal{T}}_j^k = \frac{1}{2} \Big(-\alpha P_t^0 \log \leftidx{^j}{R}{}_k^0 - (1-\alpha)P_t^1 \log \leftidx{^j}{R}{}_k^1 \Big)
%\end{equation}

\begin{equation}
    \mathcal{L}_{td} = \min_{k} \Big(-\alpha P_i^0 \log {R}{}_k^0 - (1-\alpha)P_i^1 \log {R}{}_k^1 \Big)
    \label{eq:l_td}
\end{equation}

where $\alpha$ is weight parameter used to compute the loss and the superscripts $0$ and $1$ indicates channels corresponding to traversable and non-traversable  regions of \emph{TPNet} outputs. We train the network through deep supervision by computing the \emph{Trajectory Diversity Loss} at two different levels. The first level as described above is at the last layer of the TPNet and the second level is immediately after the last decoder layer of the network. To achieve this, we use the output from the last decoder layer and apply deconvolutions on those features to produce $k \times 2$ outputs similar to the final layer of our network. The loss at this intermediate supervision is same as $\mathcal{L}_{td}$ but is applied on these intermediate outputs.

Additionally, we use an \emph{Obstacle Avoidance Loss} $\mathcal{L}_{obs}$ which penalizes the network when it predicts proposals which intersect with the obstacles in the scene. To achieve this, we minimize the negative log likelihood of $\mathcal{R}$ at obstacle locations. The loss is defined as,
\begin{equation}
    \mathcal{L}_{obs} = -\mathcal{O}^1 \log{R}{}_k^1
\end{equation}

The effective loss to train TPNet is given by 
    \begin{equation}
        \mathcal{L}_{TPN} = \mathcal{L}_{td} + \lambda\mathcal{L}_{obs}
    \end{equation}
where, $\lambda$ is the weight of \emph{Obstacle Avoidance} loss.

During the test phase, TPNet provides us with diverse trajectory regions through $\mathcal{R}$. Each of the predicted trajectory regions $R_k$ contains the probabilities associated with the pixels in trajectory regions which we further use in our next module.

\subsection{Trajectory Sampler Network}
The second module of our pipeline is the Trajectory Sampler Network which is used to predict a set of future waypoints using a particular proposal $R_k$ from TPNet. \\

\subsubsection{Architecture}
The input to the TSNet is one of the proposal, $R_k$ from the TPNet. The probability map is encoded into a feature space using a convolutional layer with kernel size of 7 followed by 3 convolutional layers of stride 2. This is fed into a series of 3 LSTM \cite{lstm} layers along with the previous predicted coordinates from the last LSTM layer. The output from the network is ${\bf \hat{W}}$ = $\{\hat{w}_1, \hat{w}_2, \hat{w}_t\}$ where $\hat{w}_t$ is a predicted trajectory coordinate. The coordinates are predicted in the discretized 2-D grid representing the scene.

\subsubsection{Training}
The ground truth coordinates are generated using an RRTstar with B-spline on top for each ${R}_k$ from the current location of the robot to a point with highest probability greater than a threshold and farthest from the robot location as the goal point. It is ensured that the eventual RRTstar trajectory lies within the trajectory proposal region output by \emph{TPNet}.  We use ${\bf W}$ = $\{w_1, w_2, ...,  w_t\}$ as the ground truth coordinates over which it is supervised. We train the network in a supervised fashion using  ${\bf \hat{W}}, {\bf W}>$ pairs. The loss function is defined as the L2 distance between the predicted and ground truth waypoint at each LSTM prediction.
% Additional info: I train with a batch size of 32 for 300 epochs
\begin{equation}
    \mathcal{L}_{TSN} = \lVert \mathbf{W} - \hat{\mathbf{W}} \rVert
\end{equation}

During test phase, each proposal $R_k$ from the TPNet is used by Trajectory Sampling network to generate diverse future trajectories for the robot.

\section{Experimental Setup}
We extensively evaluated the proposed approach on standard datasets as well as on our own University dataset. Efficacy and robustness of our approach are demonstrated by the fact that it is agnostic to the type of sensors, environments or the hardware platform chosen for experiments. This is possible as long as the sensing hardware and the input processing of point cloud is capable of providing us with sufficiently accurate measurements which can be converted to the 2D occupancy maps. We also conducted experiments to demonstrate navigation in different scenarios based on the trajectories predicted by our framework.

\subsection{Evaluation Scenarios}

\subsubsection{Standard dataset}
We demonstrate the performance of our pipeline on KITTI dataset \cite{kitti_odom}. Sequences 5,6,7,8,9,10 were used for generating the training data and the evaluation was performed on sequences 0,1,2,3,4.  
\subsubsection{Our University dataset}
We use dataset collected from our own University as an additional dataset for evaluation. In order to collect the data we use a Mahindra e2o vehicle with mounted sensors such as Velodyne-16, GPS and IMU. 

\subsection{Tests on our own setups}
Test were conducted on our platforms in two completely different scenarios. For real world outdoor experiment, we show the deployment of our pipeline on a ground robot while an aerial vehicle is used for testing in constrained practical indoor conditions.

\subsubsection{Outdoor Tests}
The outdoor tests were carried out in constrained alleys as well as on roads inside our campus. The runs in alleys were performed on a ClearPath Husky robot with Velodyne-16 mounted on top. On the University roads, a Mahindra e2o mounted with Velodyne-16 and UBlox RTK GPS with Xsens MTi-30 IMU was used as the hardware platform. In both cases a ROG GL552VX laptop with Core i7 CPU, Nvidia GTX 960M GPU and 16GB RAM was used for running the network as well as generating the control commands for trajectory tracking and navigation. The controls of steering, throttle and brakes on the car can be switched between manual mode and fully autonomous mode, where in autonomous mode we have network of sensors to implement closed loop feedback control to track the predicted trajectory.

\subsubsection{Indoor Tests}
\begin{figure*}[h]
\begin{tabular}{cccc}
\includegraphics[width=2cm, height=1.6cm]{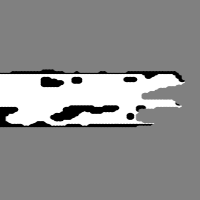} &
\includegraphics[width=2cm, height=1.6cm]{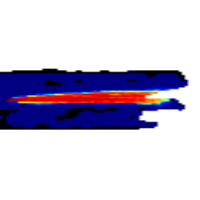} &
\includegraphics[width=2cm, height=1.6cm]{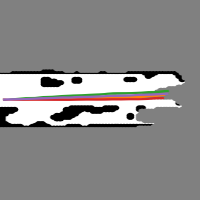} & 
\includegraphics[width=2cm, height=1.6cm]{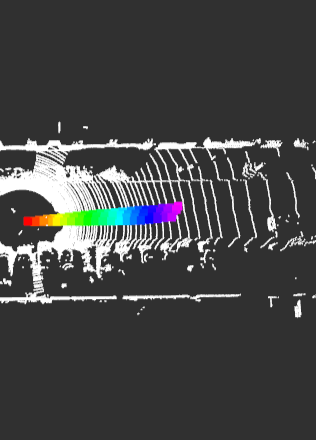} \\
\includegraphics[width=2cm, height=1.6cm]{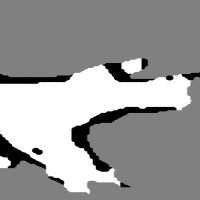} &
\includegraphics[width=2cm, height=1.6cm]{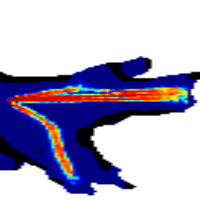} &
\includegraphics[width=2cm, height=1.6cm]{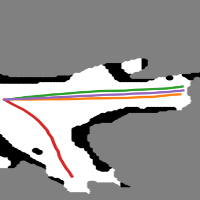} & 
\includegraphics[width=2cm, height=1.6cm]{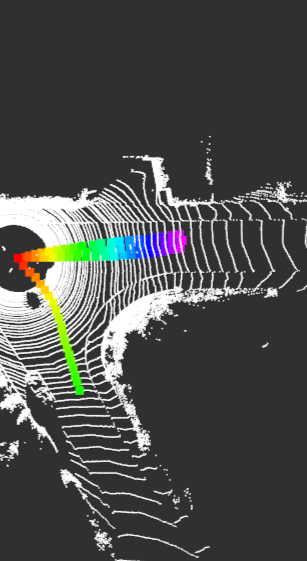} \\
\includegraphics[width=2cm, height=1.6cm]{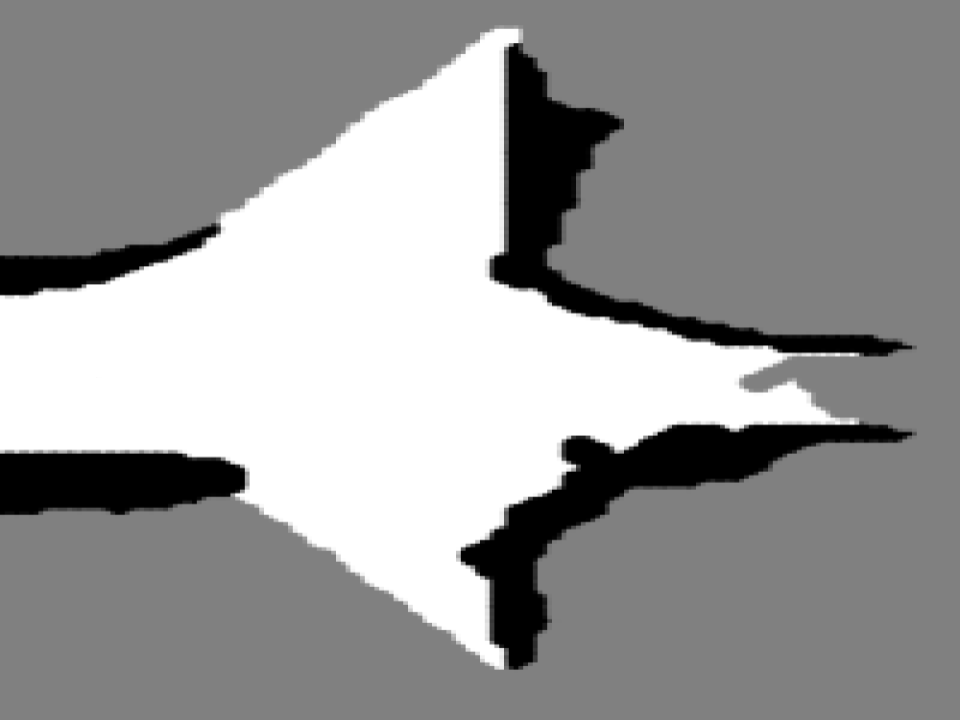} &
\includegraphics[width=2cm, height=1.6cm]{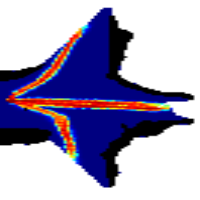} &
\includegraphics[width=2cm, height=1.6cm]{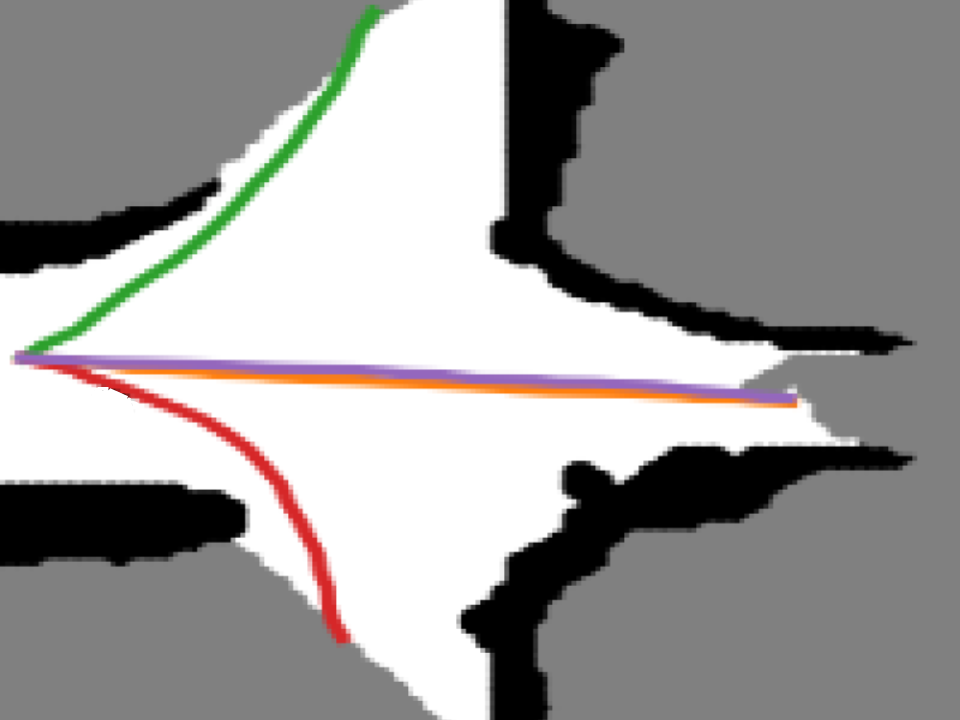} &
\includegraphics[width=2cm, height=1.6cm]{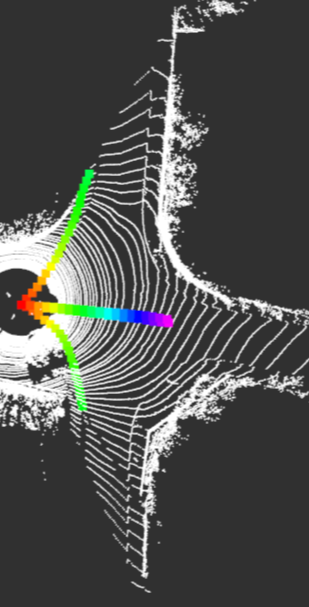} \\

\end{tabular}
\quad
\begin{tabular}{ccc}
\centering
\includegraphics[width=2cm, height=1.6cm]{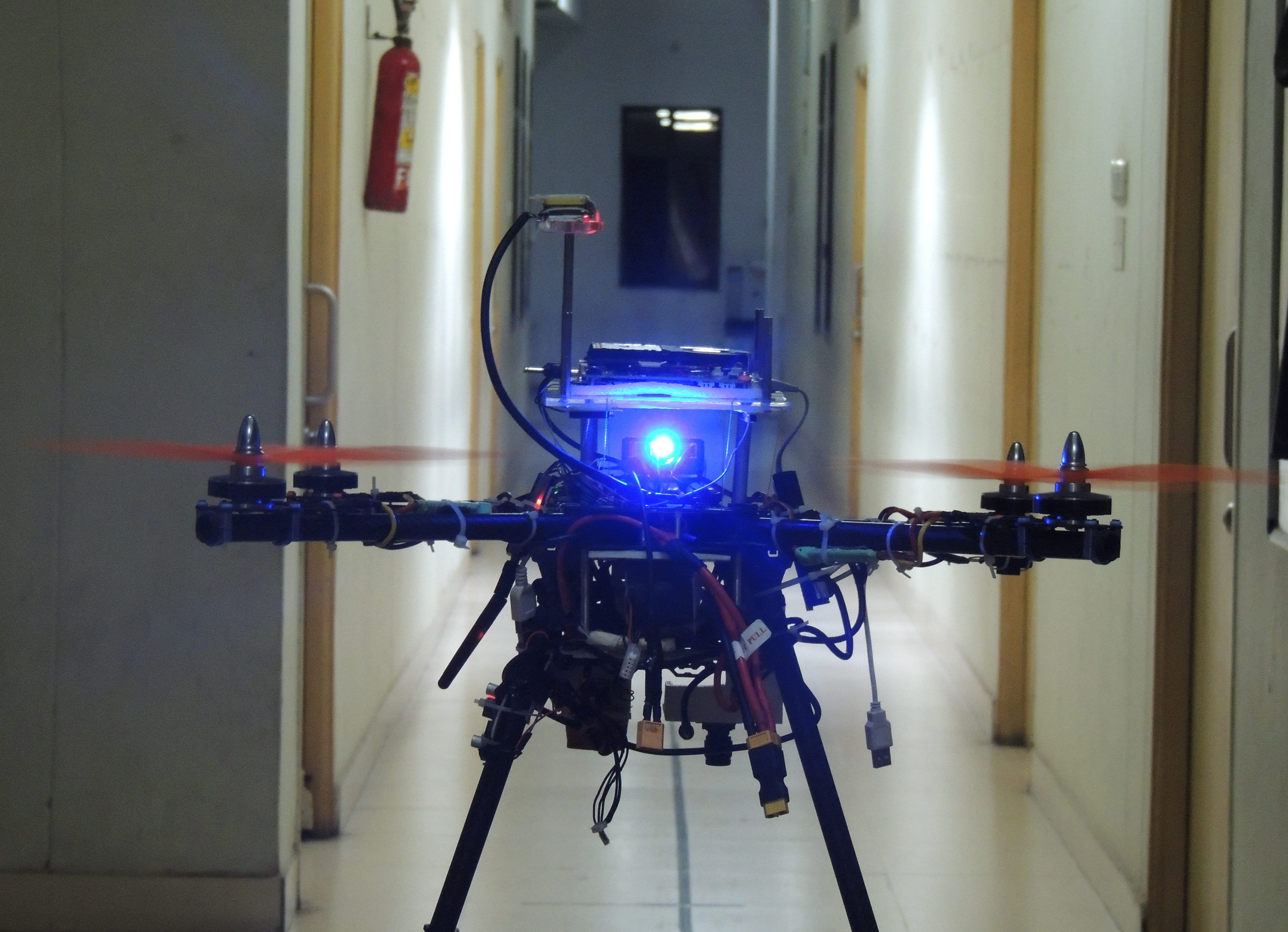}&
\includegraphics[width=2cm, height=1.6cm]{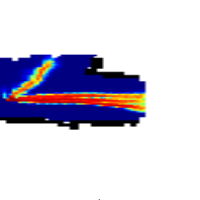}&
\includegraphics[width=2cm, height=1.6cm]{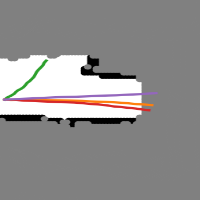}\\
\includegraphics[width=2cm, height=1.6cm]{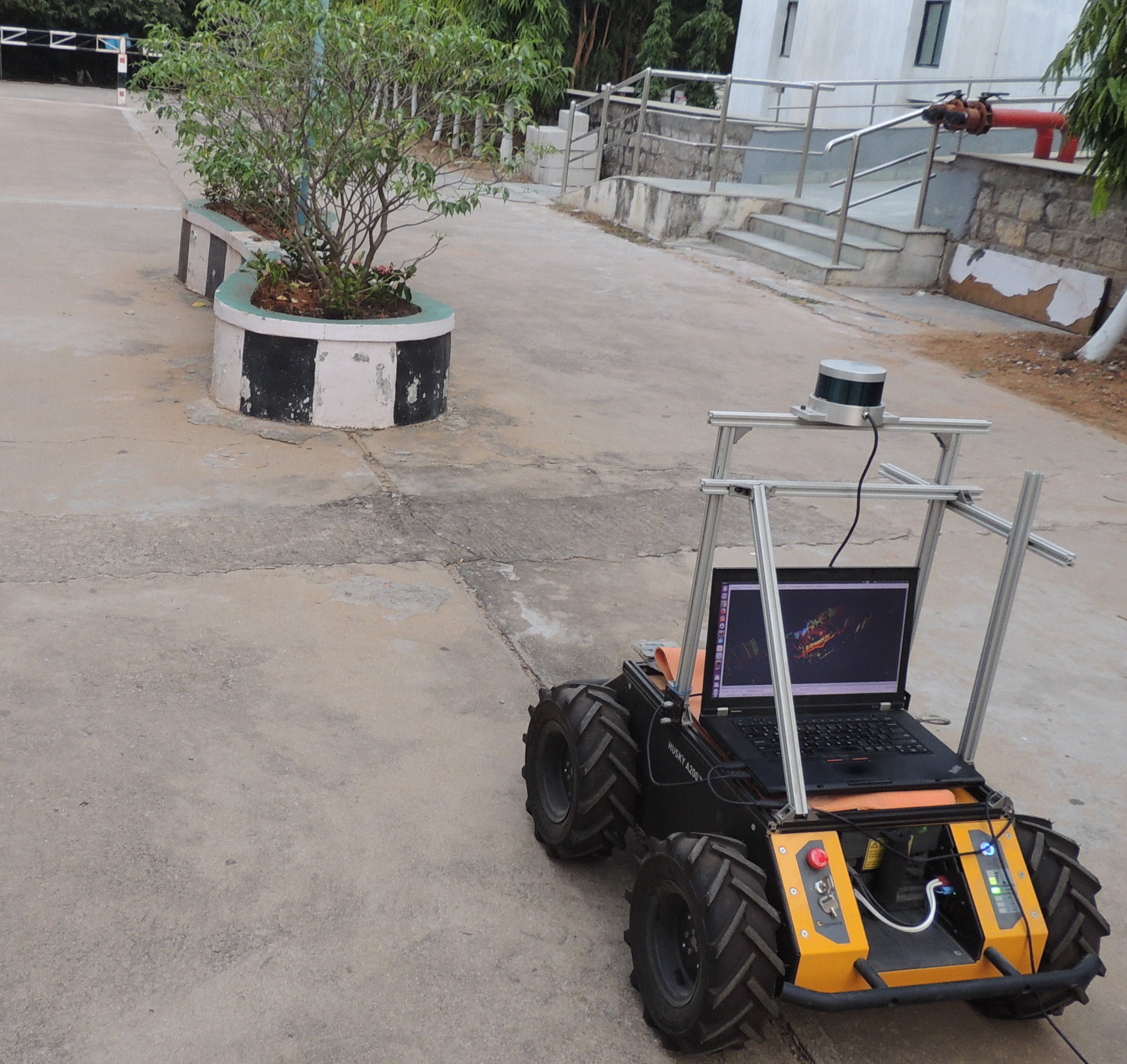}&
\includegraphics[width=2cm, height=1.6cm]{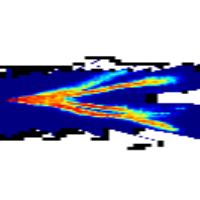} &
\includegraphics[width=2cm, height=1.6cm]{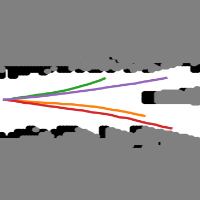} \\
\includegraphics[width=2cm, height=1.6cm]{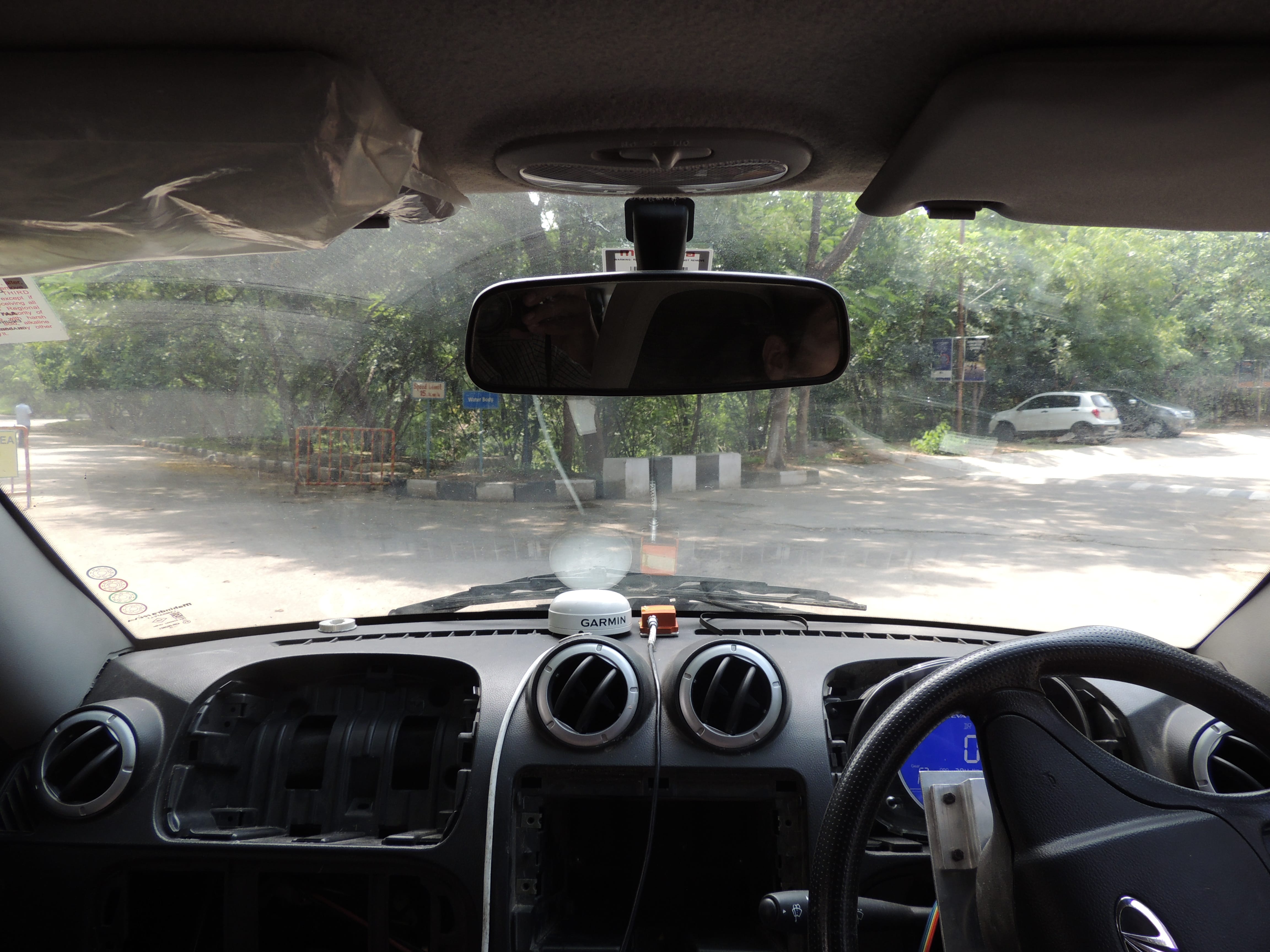}&
\includegraphics[width=2cm, height=1.6cm]{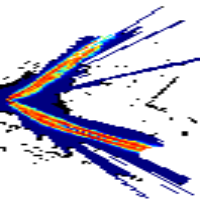} &
\includegraphics[width=2cm, height=1.6cm]{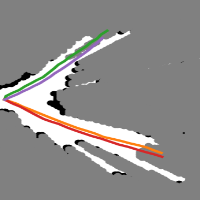} \\
\end{tabular}
\caption{{\bf Left:} KITTI results for situations with a varying number of dominant choices possible. The coloumns are represented in the order of Occupancy($\mathcal{O}$) input, prediction from TPNet, waypoints from TSNet and projected path in the pointcloud. {\bf Right:} Shows the adaptability of the network to various sensor modalities at varying situtaions such as indoors and outdoors.}
\label{fig:qualitative_results}
\end{figure*}

The indoor tests were conducted on a custom made drone(quad rotor) platform as can be seen in the figure \ref{fig:qualitative_results}. The  drone is mounted with a very low powered and light weight Intel RealSense depth sensor connected to an Intel NUC i5 processor for processing the obtained depth data. As RealSense has approx.\ang{69} horizontal field of view, the occupancy map is registered to get a static local map around the current drone position which is then passed to the ground station laptop for trajectory prediction. A monocular visual inertial odometry is running onboard for accurate state estimation. The TPNet as well as the TSNet are running on a commodity laptop mentioned above. The 2D occupancy map and the predicted trajectory is communicated between the onboard processor of the drone and the laptop over Wi-Fi. As our proposed network works with 2D occupancy maps, the trajectory is generated for a fixed height at which drone is flying.

{\it Implementation details:} The TPNet was trained using SGD optimizer with an initial learning rate of 0.005 and a batch size of 32. The learning rate was decreased every 10K steps by a factor of 0.8 and was trained for 400 epochs using  6 KITTI sequences. TSNet is also trained using SGD optimizer with an initial learning rate of 0.01 and learning rate decay of 0.7 for every 10K steps and was trained for 300 epochs. Batch size of 32 was used during the training. Both the networks were implemented in Tensorflow \cite{tensorflow2015-whitepaper}.

\begin{figure}
    \begin{tabular}{ccccc}
        \includegraphics[width=1.3cm, height=1.1cm]{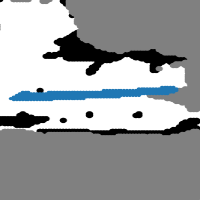}&
         \includegraphics[width=1.3cm, height=1.1cm]{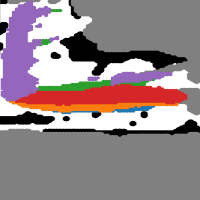}&
         \includegraphics[width=1.3cm, height=1.1cm]{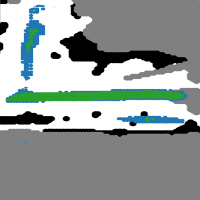} &
         \includegraphics[width=1.3cm, height=1.1cm]{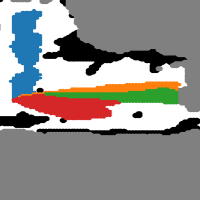} &
         \includegraphics[width=1.3cm, height=1.1cm]{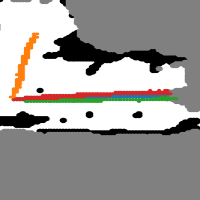}\\
         \includegraphics[width=1.3cm, height=1.1cm]{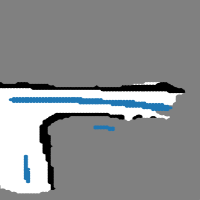}&
         \includegraphics[width=1.3cm, height=1.1cm]{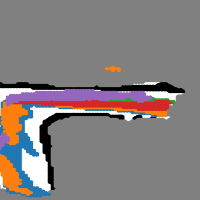}&
         \includegraphics[width=1.3cm, height=1.1cm]{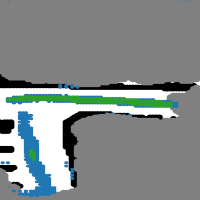} &
         \includegraphics[width=1.3cm, height=1.1cm]{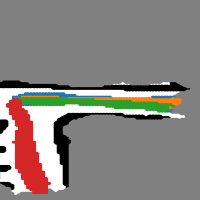} &
         \includegraphics[width=1.3cm, height=1.1cm]{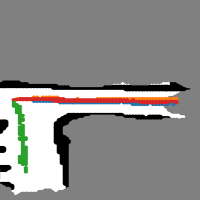}\\
         
    \end{tabular}
    \caption{Shows the ablative results for our architecture. The images shown represent the following in order (from left to right): without multiple outputs, without skip connections, without dilation, without deep supervision and with the final architecture. Points which have probabilities greater than 0.5 as a trajectory point is shown in the figure. }
    \label{fig:ablative}
\end{figure}

\begin{table*}[h!] 
  \begin{center}
    \begin{tabular}{|c|c|c|c|c|c|}
\toprule
    {\bf Datasets} & TPNet+TSNet & TPNet+RRTstar & \multicolumn{1}{c|}{RRTstar} & \multicolumn{1}{c|}{Informed-RRTstar} &\multicolumn{1}{c|}{BITstar}\\
    & & & T-PL & T-PL & T-PL\\
\midrule
    KITTI & {\bf 0.079} & 0.238 & 0.297 & 0.306 & 0.294\\
    IIIT-H & {\bf 0.082} & 0.112 & 0.192 & 0.202 & 0.190\\
    Drone & {\bf 0.079} & 0.090 & 0.147 & 0.141 & 0.139\\
    
\bottomrule
	\end{tabular}
	\caption{Compute time (in secs) for some state-of-the-art planning algorithms for calculating 4 trajectories over various datasets} \label{tab:time}
  \end{center}
  \vspace{-0.5cm}
\end{table*}

\begin{table*}[h!] 
  \begin{center}
    \begin{tabular}{|c|c|c|c|c|c|c|c|c|c|c|}
\toprule
    {\bf Datasets} & TPNet+TSNet & \multicolumn{3}{c|}{RRTstar} & \multicolumn{3}{c|}{Informed-RRTstar} &\multicolumn{3}{c|}{BITstar}\\
    & & T-PL & PL(2sec) & PL(5sec) & T-PL & PL(2sec) & PL(5sec) & T-PL & PL(2sec) & PL(5sec)\\
\midrule
    KITTI & 26.079 & 26.760 & 26.121 & 26.095 & 26.727 & 26.102 & {\bf 26.075} & 26.520 & 26.177 & 26.160\\
    IIIT-H & 24.756 & 27.427 & 24.896 & 24.757 & 27.375 & 24.803 & {\bf 24.727} & 26.239 & 24.941 & 24.813 \\
    Drone & 2.549 & 2.540 & 2.539 & 2.539 & 2.542 & 2.539 & {\bf 2.536} & 2.539 & 2.539 & 2.539 \\
    
\bottomrule
	\end{tabular}
	\caption{Path Length comparison({\it in meters}) between various planning algorithms for the same scene and goal points} \label{tab:pathlen}
  \end{center}
    \vspace{-0.8cm}
\end{table*}

\section{Results}
We evaluated the proposed approach on a wide variety of standard datasets as well as in real world scenarios. The qualitative results as well as quantitative comparisons with other works are discussed below. 

\subsection{Qualitative Results}
In figure \ref{fig:qualitative_results} we show scenarios where multiple choices of trajectories are possible. The network has an implicit understanding of the scene and predicts diverse trajectories towards various implicitly identified goal points. These are particularly the scenarios  when there are different dominant choices of motion possible (fig. \ref{fig:qualitative_results} left $2^{nd}, 3^{rd}$ row). On the other hand, when there is only one dominant choice the proposals are aligned in the same direction (fig. \ref{fig:qualitative_results} left $1^{st}$ row). It can be seen that the network predicts trajectories in different directions at complex situations such as a bifurcation or an intersection.
Although the network has been trained only on KITTI outdoor datasets, the pipeline very well generalizes for different outdoor scenes including our University driving dataset as well as to constrained indoor environments. The network was evaluated on ground robot mounted with laser scanner as well as on drone mounted with depth sensor and is found to perform equally well in all the cases (fig. \ref{fig:qualitative_results} right). The tests are primarily focused on demonstrating short term navigation based on trajectory tracking in different types of critical scenarios. 
 
{\bf Ablative study:} 
In this study, we show the importance of various components of the framework by analyzing the performance when each component is discarded. 
Firstly, we discard the diverse output $\mathcal{R}$ of TPNet and use only a single output layer. Networks with single prediction output cannot reason for multimodal possibilities of traversable areas and they tend to average out multiple of them into a single output. This can be seen in figure \ref{fig:ablative}, first column, where the network averages out all possibilities into a single one. In figure \ref{fig:ablative} second column, we show the output of the TPNet without the use of skip connections where it can be seen that the trajectories are spread across a larger area with many discontinuities. This indicates the need for skip connections the encoder-decoder architecture of TPNet to constrain its output to precise areas. In figure \ref{fig:ablative} third column, we show the output of TPNet without dilated convolutions. The proposals predicted contain significant discontinuities. This is because dilated convolutions capture spatial context of the scene much better and give out smooth transition of probabilities over the predicted proposal. In figure \ref{fig:ablative} fourth column
we show the output of TPNet when it is trained without deep supervision where the proposals are not as precise and compact as compared to our final pipeline. The computational cost increases as we supervise over many intermediate layers. Hence, we use deep supervision in our network just after the encoder-decoder. The results show that the predicted trajectories from our framework are precise and diverse as compared to previous cases.
%The computational cost increases as we supervise over many intermediate layers. Hence, we use deep supervise our network just after the encoder-decoder. 

\subsection{Quantitative Results}
We compare the proposed approach with various other state-of-the-art planning algorithms such as RRTstar \cite{rrt_star}, Informed-RRTstar \cite{informed_rrt_star} and BITstar \cite{bit_star} based on total time taken to predict same number of paths and path length given the same goal points and closest distance of the predicted path to any obstacles in the map. Moreover, we consider different optimization objectives for these planners such as Threshold-Path Length(T-PL) and Path-Length(PL), and exhibit the trade-off between time to predict a path and the path length. These results are presented in Table \ref{tab:time} and Table \ref{tab:pathlen}.
% We show a scalability plot in figure \ref{fig:sclability_plot} which depicts a near constant time taken by our network for varying number of choices predicted, while the time taken by classical planners increases by the number of outputs. It is worth mentioning that while our approach takes constant time for computation it's path length is also comparable to those predicted  by the classical ones. 

Table \ref{tab:time} shows the comparison of computation time for various algorithms such as RRTstar, Informed-RRTstar,  BITstar and TPNet+RRTstar. In order to have a fair comparison we compare the time taken by the above-mentioned planners to compute four different trajectories on same scene as input. The time for detecting goal points is also included in the total time for comparison. By default, the PL optimization runs for the max time limit specified looking for the shortest path while the T-PL is satisfied if it finds a path within a given threshold. Hence, we set the threshold path to be much higher than the maximum path length so that the planner exits as soon as it finds its first path to the goal. The values shown in the tables are the average of outputs on 1000 samples from each dataset. It is evident that proposed method consumes less time comparatively and is almost constant for various scenes. 

Table \ref{tab:pathlen} shows the trade-off between computing the optimal trajectory and the time taken to compute it.
%The trade-off between computing the optimal trajectory and that of the time taken is shown in Table\ref{tab:pathlen}. 
It represents the average path length to the goal in each of the evaluated datasets. It can be seen that the average path length reduces with increase in the max time limit for classical frameworks. While our framework takes 79 milli seconds to compute an optimal path length, Informed-RRTstar takes about 5 seconds to compute a similar optimal path length. Also, it can be noticed that our path length is very close to that of Informed-RRTstar for all the datasets. This depicts the efficiency of our framework in terms of time consumed for optimal trajectory proposal, irrespective of number of outputs. 

\section{Conclusion}
We proposed an end to end framework that maps intermediate representation in the form of an occupancy grid to multiple candidate trajectory options. These options were generated through a cascade of a Trajectory Proposal Network and a Trajectory Sampler Network. The efficacy of the network was established through its ability to discern dominant choices of motion and output trajectories that are non colliding with trajectory lengths similar to state of the art planners. But the network's ability to scale up to multiple candidate options and output multiple candidate trajectories at near constant time makes it unique and distinctive from other methods. To the best of our knowledge this is the first such architecture proposed in the literature. 

\bibliographystyle{ieeetr}
\bibliography{citations}

%\end{thebibliography}

\end{document}